\documentclass[runningheads]{llncs}

\usepackage[T1]{fontenc}
\usepackage{graphicx}

\usepackage{array}
\usepackage{booktabs}
\usepackage{multirow}

\begin{document}

\title{NeuroTS-Net: Multi-Class Semantic Segmentation of Pediatric Brain Tumors in Multi-Modal MRI}

\titlerunning{NeuroTS-Net for Pediatric Brain Tumor Segmentation} 

\author{
Darius Peteleaza\inst{1} \and
Razvan-Gabriel Dumitru\inst{2} \and
Bogdan Neamtu\inst{1,3,4} \and
Arpad Gellert\inst{1} \and
Mariana Sandu\inst{5} \and
Claudiu Matei\inst{1,5}
}

\authorrunning{D. Peteleaza et al.}

\institute{
Lucian Blaga University of Sibiu, Sibiu, Romania\\
\email{\{darius.peteleaza,bogdan.neamtu,arpad.gellert,matei.claudiu\}@ulbsibiu.ro}
\and
University of Arizona, Tucson, AZ, USA\\
\email{rdumitru@arizona.edu}
\and
Pediatric Clinical Hospital of Sibiu, Sibiu, Romania
\and
Johns Hopkins University, Baltimore, MD, USA
\and
MedLife Polisano Hospital, Sibiu, Romania\\
\email{sandu.mariana@medlife.ro}
}

\maketitle

\begin{abstract}

Pediatric brain tumors are a leading cause of cancer-related mortality in children, and their small, rare, and often low-contrast subregions make accurate manual delineation challenging. Reliable automated segmentation is therefore needed to support diagnosis, treatment planning, and response assessment. Accordingly, we introduce NeuroTS-Net, a three-dimensional encoder-decoder convolutional neural network architecture for multi-class semantic segmentation that incorporates a dual-scale raw-detail stream, adaptive low-resolution context selection, and detail-preserving multipath downsampling. These components preserve fine intensity and boundary information while efficiently modeling broader tumor context. NeuroTS-Net was trained on the BraTS 2026 pediatric dataset without external data or pretrained weights and evaluated against nnU-Net and MedNeXt under the same experimental protocol. NeuroTS-Net outperformed the baseline methods, achieving whole-tumor and tumor-core Dice scores of 0.938 and 0.937 on the internal validation set and 0.927 and 0.926 on the official challenge validation set. The code is open-sourced at: \url{https://github.com/maenstru56/NeuroTS}. 


\end{abstract}

\section{Introduction}
\label{sec:introduction}

Pediatric brain tumors are among the most common childhood cancers worldwide \cite{steliarova2017international} and remain associated with substantial mortality \cite{price2024cbtrus,girardi2023global}. Their rarity and heterogeneous presentation make multi-institutional studies and reproducible response assessment essential \cite{kazerooni2024brats}. Multiparametric magnetic resonance imaging (mpMRI) is central to diagnosis, treatment planning, and longitudinal evaluation, but manual delineation of tumor subregions is time-consuming and subject to inter-rater variability \cite{fathi2023automated}. Automated segmentation is therefore essential for consistent tumor quantification and response assessment.
 
The Brain Tumor Segmentation (BraTS) challenge has benchmarked automated segmentation methods since 2012, and in 2023 it introduced BraTS-PEDs, the first pediatric track \cite{kazerooni2024brain,kazerooni2025brain}. We use Task 2 of the BraTS 2026 cluster \cite{bakas_2026_19714728}, which targets pre-treatment and post-treatment pediatric patients and is scored on a federated platform \cite{karargyris2023federated}. Each case provides four co-registered sequences, native T1 (T1N), contrast-enhanced T1 (T1C), T2-weighted (T2W), and T2 fluid-attenuated inversion recovery (T2F), with voxel-wise labels for enhancing tumor (ET), non-enhancing tumor (NET), cystic component (CC), and peritumoral edema (ED), plus composite tumor core (TC) and whole tumor (WT).
 
Early segmentation relied on intensity thresholding and clustering methods such as k-means and fuzzy c-means \cite{negrea2026mri}, but deep learning now dominates the field. Convolutional neural networks (CNNs) such as U-Net \cite{ronneberger2015u} popularized skip-connected encoder-decoders; self-configuring nnU-Net \cite{isensee2021nnu} is a strong baseline, MedNeXt \cite{roy2023mednext,roy2025mednext} scales large kernels for volumes, and DUCK-Net \cite{dumitru2023using} uses task-specific features. For long-range context, vision transformers \cite{dosovitskiy2020image} and the Swin transformer \cite{liu2021swin} were adapted in UNETR \cite{hatamizadeh2022unetr} and Swin UNETR \cite{hatamizadeh2021swin}. State-space models such as U-Mamba \cite{ma2024u}, VM-UNet \cite{ruan2024vm}, and SegMamba \cite{xing2025segmamba} capture context efficiently, and diffusion-based methods such as MedSegDiff \cite{wu2024medsegdiff1,wu2024medsegdiff2} and ambiguity-aware formulations \cite{rahman2023ambiguous} treat segmentation as iterative denoising. Foundation models such as the Segment Anything Model \cite{kirillov2023segment} have been adapted for glioma \cite{shi2025multi}, while hybrids mix different types of architectural blocks.
 
Across BraTS-PEDs editions, the strongest solutions favor convolutional and transformer-based backbones. The first pediatric challenge was led by nnU-Net and Swin UNETR ensembles, Auto3DSeg, and self-supervised pre-training \cite{kazerooni2024brats,tang2022self}. Recent pipelines ensemble nnU-Net and MedNeXt with radiomic-guided subtyping and lesion-aware post-processing \cite{capellan2025adaptable}, refine predictions with radiologically informed cascades \cite{mulvany2025using}, and add frequency-domain decomposition \cite{shao2025rethinking} to win the 2025 challenge \cite{yi2025frequency}. Much of the gain comes from pre-processing, ensembling, and sub-region-aware post-processing once backbones saturate \cite{isensee2024nnu}. 

In this paper, we introduce NeuroTS-Net, a 3D encoder-decoder CNN for multi-class pediatric brain tumor segmentation that combines a dual-scale raw-detail stream, adaptive low-resolution context selection, and detail-preserving multipath downsampling, achieving competitive results.

\section{Methods}
\label{sec:methods}

\subsection{Model Architecture}
\label{subsec:model_architecture}

\begin{figure}[t]
    \centering
    \includegraphics[width=\textwidth]{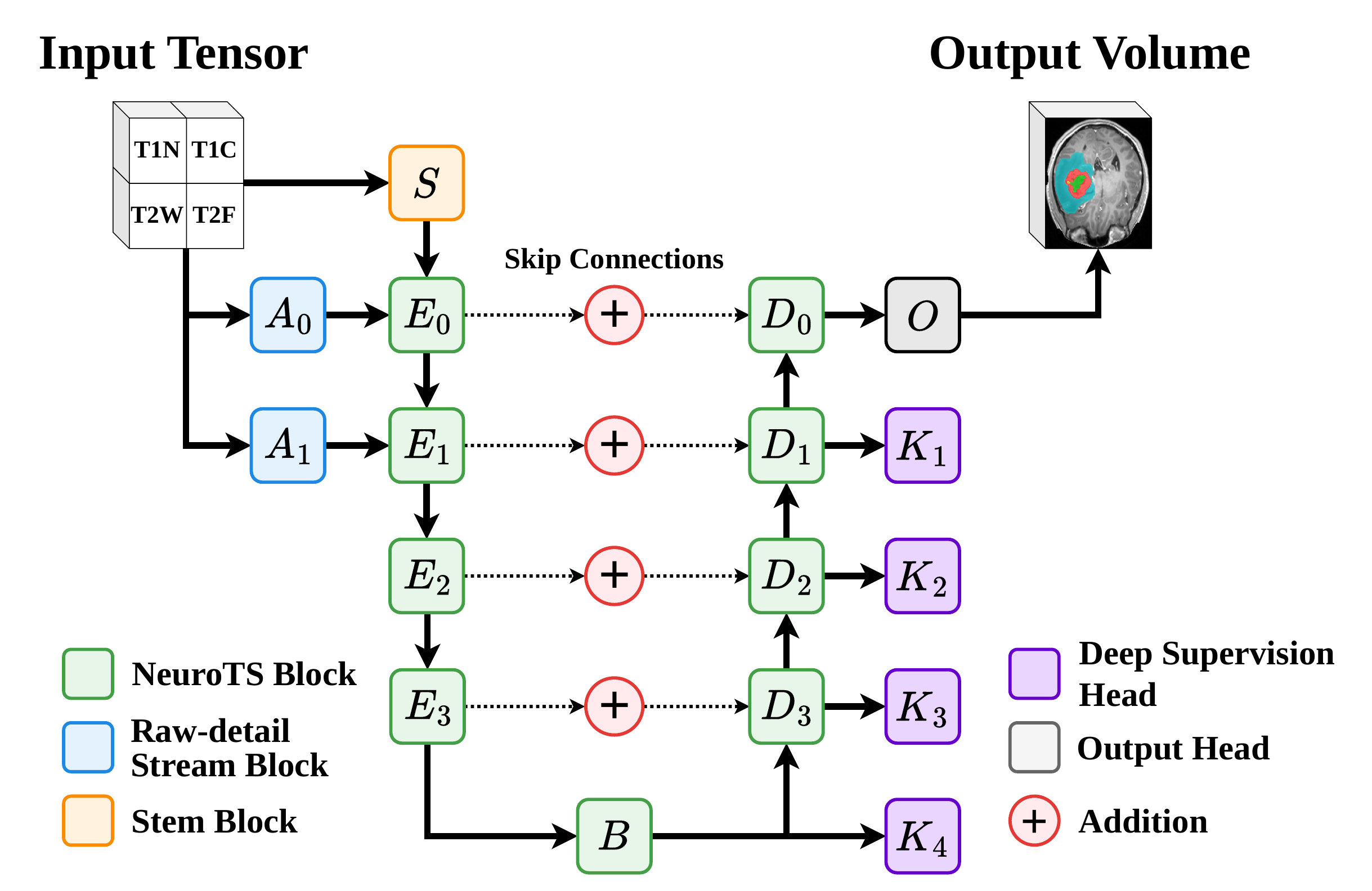}
    \caption{Overview of the proposed NeuroTS-Net architecture.}
    \label{fig:neurots_architecture}
\end{figure}

NeuroTS-Net is an encoder-decoder CNN for multi-class brain tumor segmentation. Its U-shaped hierarchy is similar to U-Net \cite{ronneberger2015u}, while MedNeXt inspires its volumetric feature-processing design \cite{roy2023mednext,roy2025mednext}. Building on these foundations, NeuroTS-Net introduces three novel components: a dual-scale raw-detail stream, adaptive context selection at low-resolution stages, and multipath downsampling that preserves both low-frequency structure and salient local responses. Together, these mechanisms improve the representation of small, low-contrast tumor regions and fine boundaries while maintaining computational efficiency.

Let \(X\in{R}^{4\times Z\times H\times W}\) denote a co-registered MRI volume containing T1N, T1C, T2W, and T2F, and let $Y\in\{0,1,2,3,4\}^{Z\times H\times W}$ denote the target segmentation, where 0 represents background and labels 1--4 represent ET, NET, CC, and ED. NeuroTS-Net learns a voxel-wise mapping from $X$ to $Y$.

As shown in Figure~\ref{fig:neurots_architecture}, the network follows a five-resolution hierarchy. The stem $S$ projects the four modalities to 32 channels, while encoder stages $E_0$--$E_3$ and bottleneck $B$ use widths of 32, 64, 128, 256, and 512 channels. The decoder $D_3$--$D_0$ mirrors the encoder and fuses each upsampled representation with the corresponding encoder feature through addition. Unlike concatenation, additive fusion preserves the decoder width and avoids additional computational costs.

The raw-detail blocks $A_0$ and $A_1$ process the input at full and half resolution. Each block combines the original intensities, a high-frequency residual obtained using $3\times3\times3$ average filtering, and the mean absolute finite-difference response along the spatial axes. These descriptors are projected to eight channels using a $1\times1\times1$ convolution and a depthwise $3\times3\times3$ convolution.

The projected detail features are multiplied by a voxel-wise sigmoid gate computed from the current encoder representation, scaled by a learnable gain initialized to $10^{-3}$, and added residually. This conditioning is applied both within the early NeuroTS blocks and after the corresponding encoder stages, refining the high-resolution skip features without dominating the initial representation.

During training, heads $K_1$--$K_4$ map $D_1$, $D_2$, $D_3$, and $B$, respectively, to auxiliary five-class logits for deep supervision. The final head $O$ applies a $1\times1\times1$ convolution to $D_0$ to produce the full-resolution logits.

\subsubsection{NeuroTS Block.}
\label{subsubsec:neurots_block}

\begin{figure}[t]
    \centering
    \includegraphics[width=\textwidth]{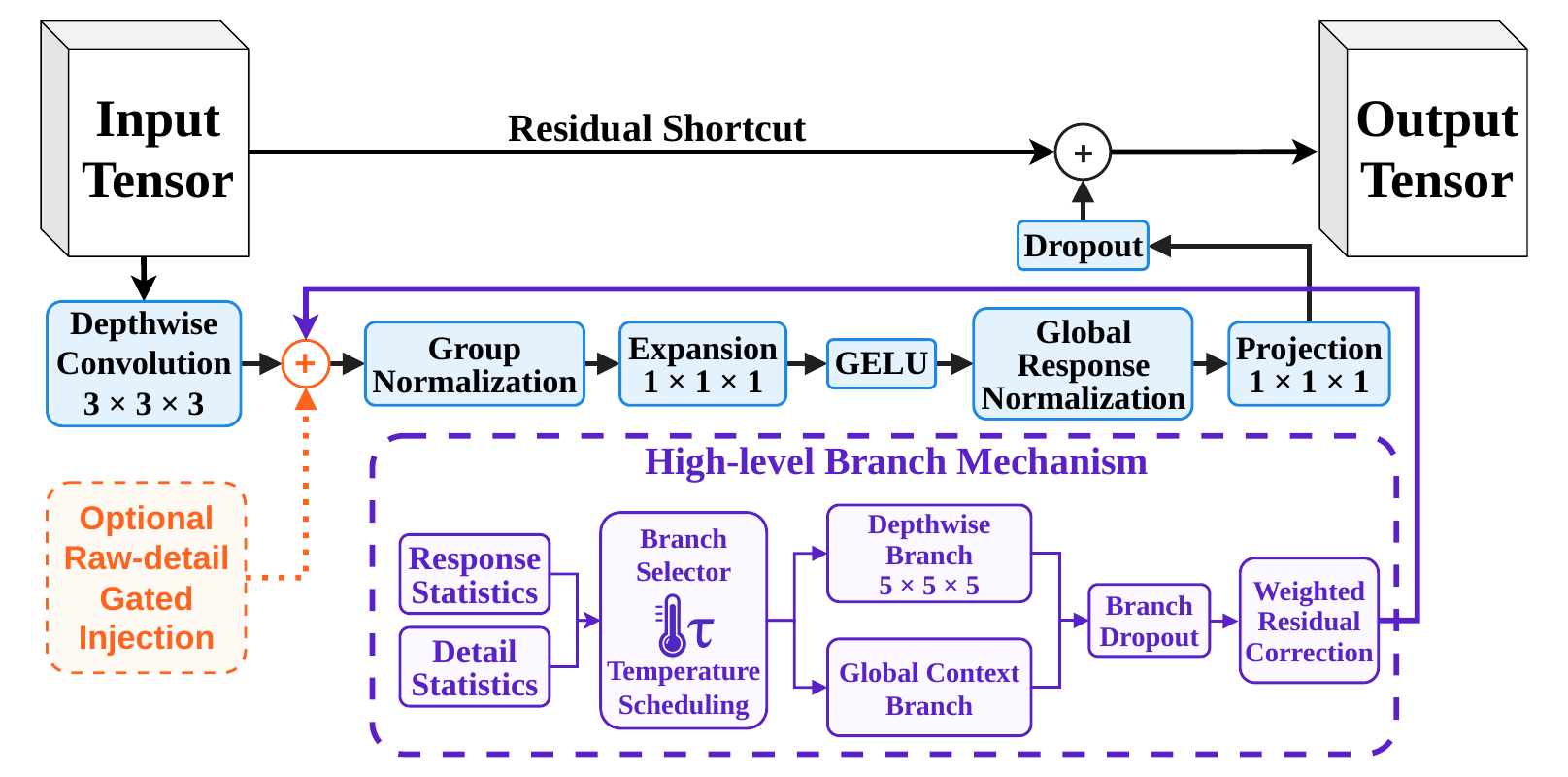}
    \caption{Overview of the proposed NeuroTS block.}
    \label{fig:neurots_block}
\end{figure}

The NeuroTS block, shown in Figure~\ref{fig:neurots_block}, is inspired by the design of MedNeXt \cite{roy2023mednext,roy2025mednext} and the adaptive routing principle of selective kernel networks \cite{li2019selective}. It extends these foundations with two novel mechanisms: gated raw-detail conditioning at high-resolution stages and response- and detail-conditioned context selection at low-resolution stages. Together, these additions preserve subtle boundary and intensity information while adaptively increasing the receptive field according to the input, feature channel, and spatial location.

At encoder stages 0 and 1, gated raw-detail injection repeatedly supplies intensity, edge, and gradient information to the early blocks and high-resolution skip features. At lower resolutions, the local depthwise response is supplemented by a depthwise $5\times5\times5$ branch at stage 3 and an additional global-context branch at the bottleneck, implemented using global average pooling, channel-wise projection, and spatial broadcasting.

The selector combines responses normalized across channels and branches with a local high-frequency signal, then converts the resulting scores and learned priors into temperature-scaled routing weights. Its gain is initialized to $0.005$, branch priors are zero-initialized, and the temperature decreases linearly from $2.0$ to $0.5$. A branch dropout of $0.05$ and early regularization discourage branch collapse. The result is processed by Group Normalization, expansion, Gaussian Error Linear Unit (GELU) \cite{hendrycks2016gaussian}, Global Response Normalization, and projection before residual addition.

\subsubsection{Downsampling Mechanism.}
\label{subsubsec:downsampling}

\begin{figure}[t]
    \centering
    \includegraphics[width=\textwidth]{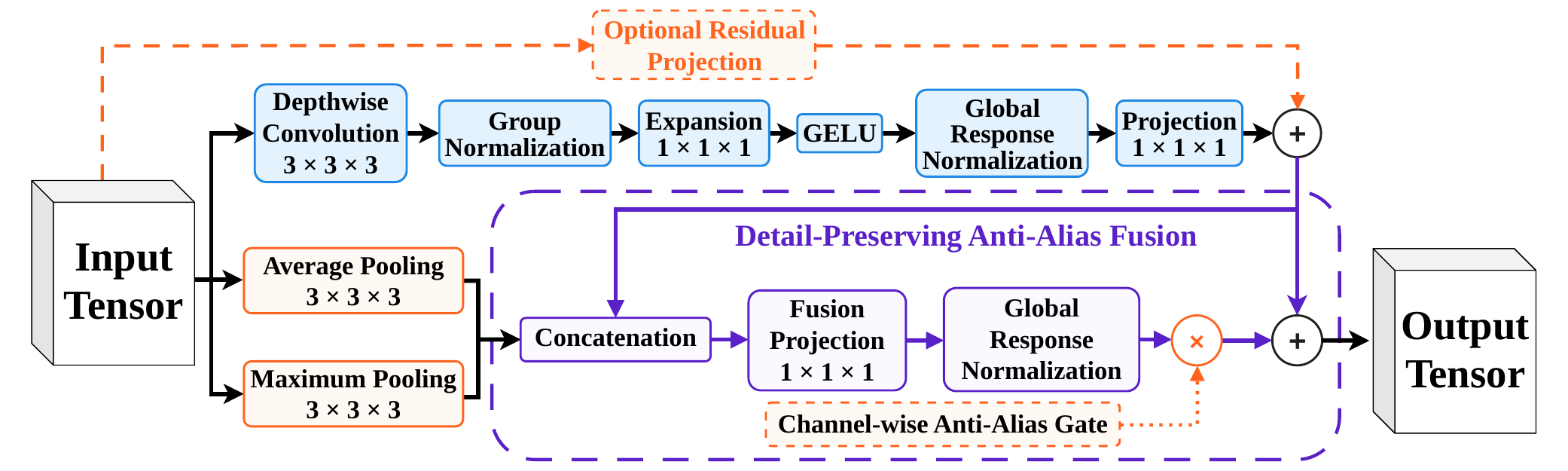}
    \caption{The proposed detail-preserving multipath downsampling mechanism.}
    \label{fig:neurots_downsampling}
\end{figure}

NeuroTS introduces a detail-preserving multipath downsampling mechanism that augments the learned stride-two transition. This mechanism is used in all encoder transitions. As illustrated in Figure~\ref{fig:neurots_downsampling}, average pooling provides a low-pass representation that reduces aliasing, while maximum pooling preserves salient local responses that may correspond to thin boundaries or small tumor components.

The learned response is concatenated with the average- and maximum-pooled representations, then compressed to the target channel width using a $1\times1\times1$ projection and refined by Global Response Normalization. The resulting correction is added to the learned path through a channel-wise gate initialized to zero. Thus, each transition initially behaves exactly like the original learned downsampling operation, while training can progressively introduce positive or negative low-pass and detail-preserving corrections for individual output channels.

\subsection{Dataset Description}
\label{subsec:dataset_description}

We use the dataset released for Task 2 of the BraTS 2026 pediatric brain tumor segmentation challenge \cite{bakas_2026_19714728}. No external data or pretrained models were used. Each case contains T1N, T1C, T2W, and T2F volumes with voxel-wise annotations for ET, NET, CC, and ED, encoded as labels 1--4. The composite TC comprises ET, NET, and CC, while the WT includes all foreground classes \cite{kazerooni2025brain}.

The two training releases contained 294 annotated volumes after deduplication. We created a patient-grouped split of 254 training and 40 internal validation volumes. The validation cohort was selected to represent challenging tissue configurations and contained 22 ET-positive, 16 CC-positive, 14 ED-positive, and 14 difficult ED-negative cases. WT and TC volumes were balanced across four volume quantiles, with additional representation of 7 small ET, 7 small CC, and 6 small ED cases. The official validation cohort contained 91 cases without public annotations and was used exclusively for evaluation on the Synapse platform.

\subsection{Data Preprocessing}
\label{subsec:preprocessing}

Volumes were retained in the challenge-provided geometry and cropped to the foreground bounding box, expanded by 32 voxels and clipped to the image extent.

Intensity normalization was applied separately to each modality ($c$). Intensities were clipped to the 0.5th--99.5th percentile range and standardized using the mean ($\mu_c$) and standard deviation ($\sigma_c$)  of the remaining nonzero voxels:
\begin{equation}
I_c^{\mathrm{norm}}
=
\frac{{clip}(I_c,q_{0.5,c},q_{99.5,c})-\mu_c}{\sigma_c},
\end{equation}
where $q_{0.5,c}$ and $q_{99.5,c}$ are the modality-specific percentile bounds. 

\subsection{Data Augmentation}
\label{subsec:data_augmentation}

Spatial and intensity augmentations were applied online during training to all four modalities and the corresponding segmentation map \cite{shorten2019survey}. Spatial transformations included random rotations of up to $\pm30^{\circ}$ ($p=0.2$), isotropic scaling in $[0.8,1.25]$ ($p=0.2$), and independent mirroring along each spatial axis ($p=0.5$). 

Intensity augmentation comprised Gaussian noise with variance in $[0,0.1]$ ($p=0.1$), Gaussian blur with standard deviation in $[0.5,1.0]$ ($p=0.15$), brightness and contrast adjustment ($p=0.15$), and low-resolution simulation ($p=0.1$). Gamma transformations used $\gamma\in[0.7,1.5]$, with probabilities of $0.3$ for the standard variant and $0.1$ for intensity inversion followed by gamma correction.

\subsection{Model Training}
\label{subsec:training}

NeuroTS-Net was trained from scratch for 1000 epochs using patches of size $128\times160\times112$, a batch size of 4, and 250 iterations per epoch. Patch types were sampled as random, foreground, ET, NET, CC, ED, and hard negative with probabilities $0.15$, $0.20$, $0.15$, $0.10$, $0.20$, $0.12$, and $0.08$, respectively. For class-targeted patches, a positive case, one of its 26-connected components, and a voxel within that component were sampled uniformly. Hard negatives were primarily drawn from ED-absent cases and centered on non-ED tumor tissue.

Training minimized the objective in Eq.~\ref{eq:training_loss}, which combines weighted cross-entropy ($\mathcal{L}_{\mathrm{CE}}$), foreground Dice ($\mathcal{L}_{\mathrm{Dice}}$), region overlap ($\mathcal{L}_{\mathrm{Region}}$), and an absent-class false-positive penalty ($\mathcal{L}_{\mathrm{Abs}}$):
\begin{equation}
\mathcal{L}
=
0.40\mathcal{L}_{\mathrm{CE}}
+0.40\mathcal{L}_{\mathrm{Dice}}
+0.10\mathcal{L}_{\mathrm{Region}}
+0.10\mathcal{L}_{\mathrm{Abs}}.
\label{eq:training_loss}
\end{equation}
The CE weights for background, ET, NET, CC, and ED were $0.05$, $2.25$, $1.00$, $3.50$, and $2.00$, while Dice weights were $1.40$, $0.60$, $2.20$, and $1.40$. The region term used Tversky overlap \cite{salehi2017tversky} for ET, CC, and ED, with $(\alpha,\beta)$ set to $(0.45,0.55)$, $(0.50,0.50)$, and $(0.60,0.40)$, respectively, and binary Dice for NET, TC, and WT. $\mathcal{L}_{\mathrm{Abs}}$ penalized false-positive predictions by averaging their squared predicted probabilities over ET, CC, and ED classes absent from the ground truth.

Optimization used AdamW \cite{kingma2014adam,loshchilov2017decoupled} with a learning rate of $1.2\times10^{-3}$, weight decay $3\times10^{-5}$, and a 10-epoch linear warmup followed by cosine decay to $10^{-6}$ \cite{loshchilov2016sgdr}. Gradients were clipped to a norm of 12, and the training was implemented in PyTorch \cite{paszke2019pytorch}, also using automatic mixed precision \cite{micikevicius2017mixed}. The branch-entropy regularizer had weight $10^{-4}$ and was active during the first 30\% of training.

Training was performed on a server with an NVIDIA H100 94GB GPU, an Intel Xeon Gold 6526Y CPU, and 128GB DDR5 RAM.

\begin{table}[t]
\caption{Comparison of quantitative lesion-wise Dice results across the BraTS-PEDs training, internal validation (INT VAL), and official validation (PED VAL) sets.}
\label{tab:dice-results}
\centering
\setlength{\tabcolsep}{5pt}
\renewcommand{\arraystretch}{1.18}
\resizebox{\textwidth}{!}{%
\begin{tabular}{
@{}
>{\centering\arraybackslash}m{1.35cm}
@{\hspace{0pt}}
>{\centering\arraybackslash}m{3.5cm}
@{\hspace{0pt}}
*{6}{c}
@{}
}
\toprule
\multirow{2}{*}{\textbf{Task}} &
\multirow{2}{*}{\textbf{Method}} &
\multicolumn{6}{c}{\textbf{Dice $\uparrow$}} \\
\cmidrule(lr){3-8}
& &
\textbf{WT} & \textbf{TC} & \textbf{NET} &
\textbf{ET} & \textbf{CC} & \textbf{ED} \\
\midrule
\shortstack[c]{%
\textbf{TRAIN}\\
\textbf{$N=254$}%
}
& \textbf{NeuroTS-Net}
& \textbf{0.942} & \textbf{0.941} & \textbf{0.906}
& \textbf{0.845} & \textbf{0.777} & \textbf{0.826} \\
\midrule
\multirow{4}{*}{%
\shortstack[c]{%
\textbf{INT}\\
\textbf{VAL}\\
\textbf{$N=40$}%
}%
}
& nnU-Net (XL)
& 0.920 & 0.919 & 0.889 & 0.575 & 0.499 & 0.521 \\
& nnU-Net ResEnc (XL)
& 0.922 & 0.918 & 0.891 & 0.596 & 0.538 & 0.530 \\
& MedNeXt (L)
& 0.925 & 0.924 & 0.888 & 0.598 & 0.559 & 0.576 \\
& \textbf{NeuroTS-Net}
& \textbf{0.938} & \textbf{0.937} & \textbf{0.901}
& \textbf{0.623} & \textbf{0.575} & \textbf{0.590} \\
\midrule
\multirow{4}{*}{%
\shortstack[c]{%
\textbf{PED}\\
\textbf{VAL}\\
\textbf{$N=91$}%
}%
}
& nnU-Net (XL)
& 0.917 & 0.918 & 0.888 & 0.444 & 0.128 & 0.000 \\
& nnU-Net ResEnc (XL)
& 0.918 & 0.916 & 0.881 & 0.463 & 0.169 & 0.000 \\
& MedNeXt (L)
& 0.922 & 0.922 & 0.894 & 0.447 & 0.140 & 0.000 \\
& \textbf{NeuroTS-Net}
& \textbf{0.927} & \textbf{0.926} & \textbf{0.895}
& \textbf{0.518} & \textbf{0.231} & \textbf{0.000} \\
\bottomrule
\end{tabular}%
}
\end{table}

\subsection{Data Post-processing}
\label{subsec:postprocessing}

Connected-component filtering was applied only to ET and CC predictions. ET logits were increased by $0.25$, thresholded at $0.30$, and restricted to components of at least 20 voxels, with at most four retained. CC used a logit bias of $0.75$, a probability threshold of $0.15$, the same minimum component size, and at most three retained components. TC and WT were preserved during reassignment.

\subsection{Evaluation}
\label{subsec:evaluation}

We report lesion-wise Dice and 95th-percentile Hausdorff distance (HD95) for ET, NET, CC, ED, TC, and WT. Higher Dice and lower HD95 indicate better volumetric overlap and boundary agreement, respectively. Comparisons include nnU-Net \cite{isensee2021nnu} and MedNeXt \cite{roy2023mednext,roy2025mednext}, each trained under the same data split and experimental protocol as NeuroTS-Net. Their configurations were determined by the nnU-Net planner, and both were implemented within the nnU-Net framework, whereas NeuroTS-Net used our independent framework. In accordance with the challenge rules, no external data or pretrained models were used.

\section{Results}
\label{sec:results}

\subsection{Quantitative Results}
\label{subsec:results_quantitative}

Table~\ref{tab:dice-results} shows that NeuroTS-Net achieved the highest Dice for all six regions on the internal validation set and for five regions on the official validation set. The largest gains were observed for ET and CC, while ED remained challenging on the official validation set, where all compared methods received a Dice of zero. 

Table~\ref{tab:hd95-results} shows that NeuroTS-Net achieved the lowest HD95 for five regions on both the internal and official validation sets, while nnU-Net ResEnc and MedNeXt achieved the lowest CC HD95 on the internal and official sets, respectively.

NeuroTS-Net is the most compact model, with 18.7 million (M) parameters, compared with 31.2M for nnU-Net, 102.4M for nnU-Net ResEnc, and 62.9M for MedNeXt. It also achieved the fastest mean inference time per case of 32.7 seconds, compared with approximately 37.7, 41.3, and 52.7 seconds, respectively.

\begin{table}[t]
\caption{Comparison of quantitative lesion-wise HD95 results across the BraTS-PEDs training, internal validation (INT VAL), and official validation (PED VAL) sets.}
\label{tab:hd95-results}
\centering
\setlength{\tabcolsep}{5pt}
\renewcommand{\arraystretch}{1.18}
\resizebox{\textwidth}{!}{%
\begin{tabular}{
@{}
>{\centering\arraybackslash}m{1.35cm}
@{\hspace{0pt}}
>{\centering\arraybackslash}m{3.5cm}
@{\hspace{0pt}}
*{6}{c}
@{}
}
\toprule
\multirow{2}{*}{\textbf{Task}} &
\multirow{2}{*}{\textbf{Method}} &
\multicolumn{6}{c}{\textbf{HD95 (mm) $\downarrow$}} \\
\cmidrule(lr){3-8}
& &
\textbf{WT} & \textbf{TC} & \textbf{NET} &
\textbf{ET} & \textbf{CC} & \textbf{ED} \\
\midrule
\shortstack[c]{%
\textbf{TRAIN}\\
\textbf{$N=254$}%
}
& \textbf{NeuroTS-Net}
& \textbf{2.045} & \textbf{1.978} & \textbf{2.085}
& \textbf{1.868} & \textbf{3.056} & \textbf{2.497} \\
\midrule
\multirow{4}{*}{%
\shortstack[c]{%
\textbf{INT}\\
\textbf{VAL}\\
\textbf{$N=40$}%
}%
}
& nnU-Net (XL)
& 9.842 & 9.681 & 9.896 & 6.621 & 4.253 & 7.612 \\
& nnU-Net ResEnc (XL)
& 8.692 & 8.703 & 8.923 & 6.401 & \textbf{3.431} & 6.286 \\
& MedNeXt (L)
& 5.011 & 3.995 & 4.617 & 4.274 & 3.836 & 6.771 \\
& \textbf{NeuroTS-Net}
& \textbf{4.092} & \textbf{3.690} & \textbf{4.289}
& \textbf{4.157} & 3.694 & \textbf{5.887} \\
\midrule
\multirow{4}{*}{%
\shortstack[c]{%
\textbf{PED}\\
\textbf{VAL}\\
\textbf{$N=91$}%
}%
}
& nnU-Net (XL)
& 9.535 & 9.029 & 9.016 & 122.8 & 222.6 & 373.0 \\
& nnU-Net ResEnc (XL)
& 8.464 & 8.017 & 8.927 & 130.9 & 271.1 & 373.0 \\
& MedNeXt (L)
& 5.981 & 4.772 & 5.472 & 129.9 & \textbf{221.1} & 373.0 \\
& \textbf{NeuroTS-Net}
& \textbf{3.610} & \textbf{3.609} & \textbf{4.376}
& \textbf{119.6} & 233.4 & \textbf{373.0} \\
\bottomrule
\end{tabular}%
}
\end{table}

\begin{figure}[!ht]
    \centering
    \includegraphics[width=\textwidth]{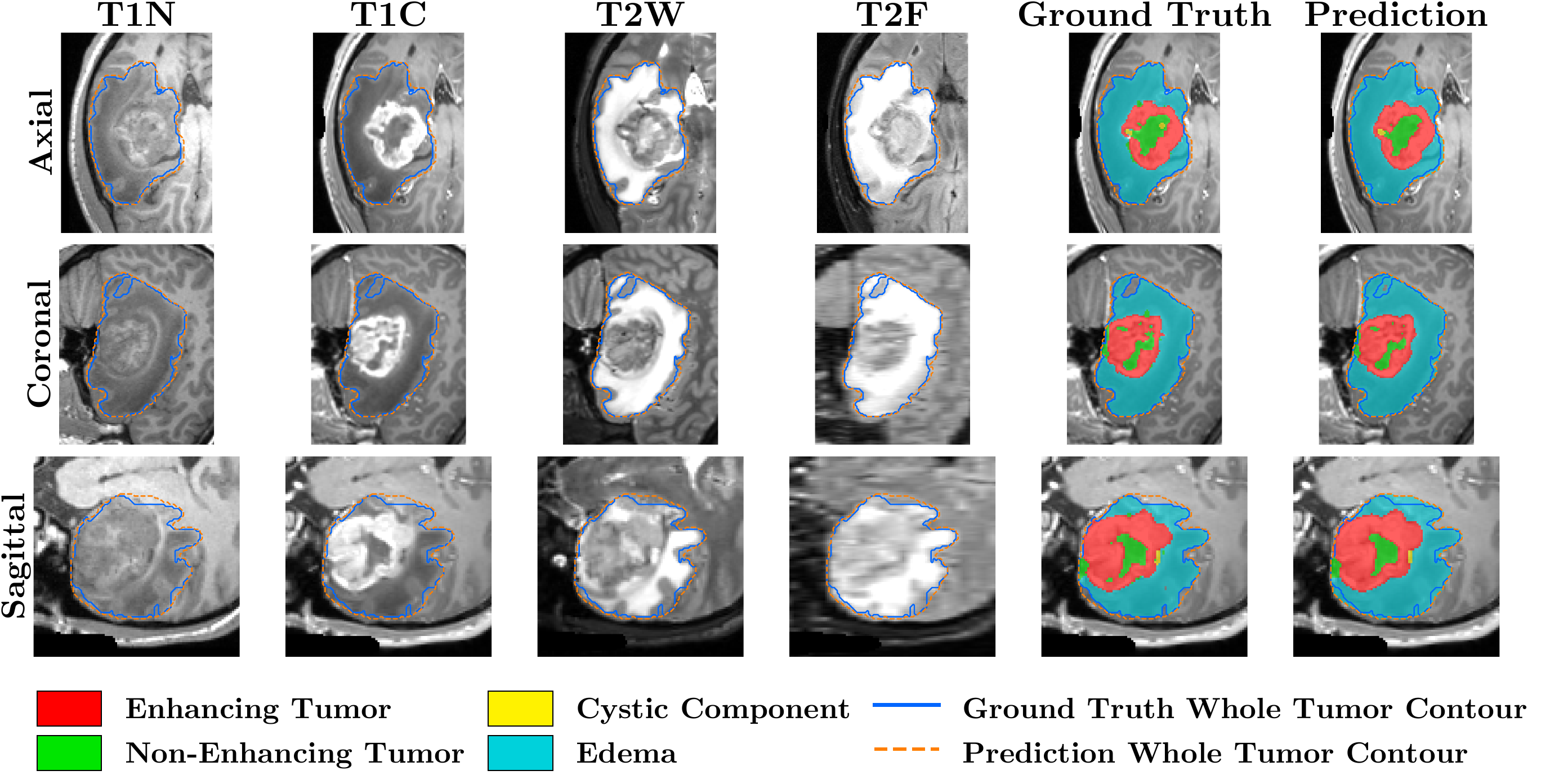}
    \caption{Qualitative comparison of a NeuroTS-Net prediction with the ground-truth segmentation in the axial, coronal, and sagittal planes.}
    \label{fig:comparison-2d}
\end{figure}

\subsection{Qualitative Results}
\label{subsec:results_qualitative}

\begin{figure}[t]
    \centering
    \includegraphics[width=0.95\textwidth]{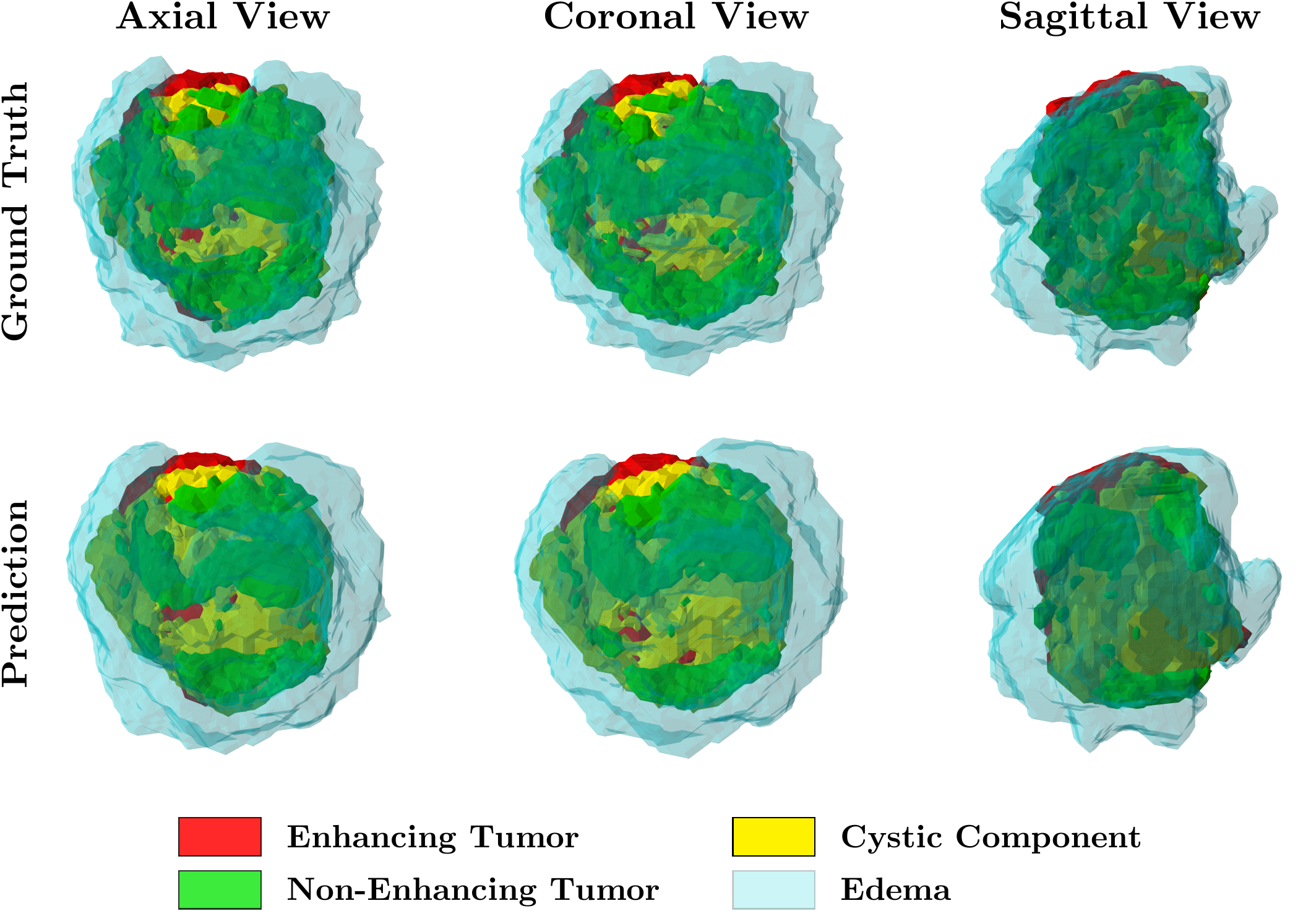}
    \caption{Three-dimensional visualization of the ground-truth segmentation and NeuroTS-Net prediction from axial, coronal, and sagittal viewpoints.}
    \label{fig:comparison-3d}
\end{figure}

Figure~\ref{fig:comparison-2d} presents a case containing all four tissue classes across the axial, coronal, and sagittal planes. The predicted WT contour closely follows the reference, with only minor deviations at peripheral regions. ET and NET are reproduced consistently, while CC and ED show small local boundary differences. The complementary MRI information is also evident, with T1C emphasizing enhancing tissue and T2W and T2F highlighting fluid-rich regions and edema.

The 3D comparison in Figure~\ref{fig:comparison-3d} further shows close spatial agreement between the ground truth and prediction across all three viewpoints. NeuroTS-Net preserves the overall tumor geometry, including the central tumor components and surrounding edema, with differences mainly confined to local class boundaries.

\subsection{Ablation of NeuroTS-Net Architectural Components}
\label{subsec:results_ablation}

Table~\ref{tab:ablation} shows that all proposed components contribute to performance across the evaluated regions. Removing the raw-detail stream, context selection, or multipath downsampling consistently reduced Dice compared with the complete NeuroTS-Net. The largest reductions were observed for the smaller tumor regions, particularly ET, CC, and ED, while WT and TC were less affected. These results indicate that the proposed components provide complementary benefits, with the full architecture achieving the highest Dice across all six regions.

\begin{table}[t]
\caption{Ablation study of the NeuroTS-Net components on the internal validation (INT VAL) set. Without (w/o) indicates removal of the corresponding component.}
\label{tab:ablation}
\centering
\setlength{\tabcolsep}{5pt}
\renewcommand{\arraystretch}{1.18}
\resizebox{\textwidth}{!}{%
\begin{tabular}{
@{}
>{\centering\arraybackslash}m{1.35cm}
@{\hspace{0pt}}
>{\centering\arraybackslash}m{3.5cm}
@{\hspace{0pt}}
*{6}{c}
@{}
}
\toprule
\multirow{2}{*}{\textbf{Task}} &
\multirow{2}{*}{\textbf{Variant}} &
\multicolumn{6}{c}{\textbf{Dice $\uparrow$}} \\
\cmidrule(lr){3-8}
& &
\textbf{WT} & \textbf{TC} & \textbf{NET} &
\textbf{ET} & \textbf{CC} & \textbf{ED} \\
\midrule
\multirow{4}{*}{%
\shortstack[c]{%
\textbf{INT}\\
\textbf{VAL}\\
\textbf{$N=40$}%
}%
}
& w/o Raw-Detail Stream
& 0.926 & 0.925 & 0.889 & 0.595 & 0.548 & 0.564 \\

& w/o Context Selection
& 0.924 & 0.922 & 0.863 & 0.599 & 0.551 & 0.568 \\

& w/o Downsampling
& 0.930 & 0.928 & 0.888 & 0.593 & 0.552 & 0.569 \\

& \textbf{NeuroTS-Net}
& \textbf{0.938} & \textbf{0.937} & \textbf{0.901}
& \textbf{0.623} & \textbf{0.575} & \textbf{0.590} \\
\bottomrule
\end{tabular}%
}
\end{table}

\section{Discussion}
\label{sec:discussion}

We introduced NeuroTS-Net, a 3D pediatric brain tumor segmentation network combining multipath downsampling, adaptively selected low-resolution context, and dual-scale raw-detail conditioning. NeuroTS-Net outperformed the compared methods, achieving WT and TC Dice scores of $0.938$ and $0.937$ on the internal validation set and $0.927$ and $0.926$ on the official validation set. Across all six regions, the mean Dice was $0.873$ on the training set, $0.761$ on the internal validation set, and $0.583$ on the official validation set. The ablation study further showed consistent performance reductions when removing any proposed component, supporting their complementary contribution. Raw-detail conditioning and multipath downsampling particularly benefited ET, CC, and ED, while context selection produced the largest NET reduction when removed.

The main limitation is reduced robustness for rare classes, particularly ED and CC. Their performance decreased on the official validation set, which may reflect rare-class overrepresentation in the internal split and cohort differences in class prevalence or imaging characteristics. The official lesion-wise protocol further amplifies ED detection errors: when no predicted and reference ED components are matched, Dice is set to zero, and HD95 receives the maximum penalty. Thus, missed ED or false positives in ED-absent cases can strongly affect the score, with all compared methods receiving the same ED penalty on the official validation set. Evaluation on a single internal split and training seed also limits assessment of patient and training variability.

Further work should improve ED and CC refinement, class-presence calibration, and boundary-aware objectives. Evaluation on additional patient-grouped splits and random seeds would further assess robustness and generalization.

\begin{credits}

\subsubsection{\ackname}
This work was supported by the Swiss National Science Foundation and UEFISCDI under the Second Swiss Contribution through the Multilateral Academic Projects (MAPS) project ``AI-based Brain Metastases Tracking and Segmentation (A-BEACON),'' grant no.~IZ11Z0\_230215, contract UEFISCDI no.~14ROCH\allowbreak/2025, and project code F-RO-CH-2024-0233, for the 2025--2029 project period.

\subsubsection{\discintname}
The authors declare no competing interests.

\end{credits}

\bibliographystyle{splncs04}
\bibliography{references}

\end{document}